\documentclass[nohyperref]{article}

\usepackage{microtype}
\usepackage{graphicx}
\usepackage{subfigure}
\usepackage{booktabs} 
\usepackage{times}
\usepackage{latexsym}
\usepackage{graphicx}
\usepackage{booktabs}
\usepackage{makecell}
\usepackage{amsmath}
\usepackage{graphicx}
\usepackage{tablefootnote}
\usepackage{amssymb}
\usepackage[T1]{fontenc}
\usepackage{xspace}
\usepackage[utf8]{inputenc}

\usepackage{microtype}

\usepackage{amsmath}
\usepackage{xcolor}
\usepackage{inconsolata}

\usepackage{hyperref}

\newcommand*{\xdash}[1][3em]{\rule{1cm}{0.15mm}{}}

\newcommand{\model}{\textsc{RePlug}\xspace}
\newcommand{\tunemodel}{\textsc{RePlug LSR}\xspace}

\usepackage[accepted]{icml2022}

\usepackage{amsmath}
\usepackage{tabularx}

\usepackage{amssymb}
\usepackage{mathtools}
\usepackage{amsthm}

\usepackage[capitalize,noabbrev]{cleveref}

\theoremstyle{plain}

\theoremstyle{definition}

\theoremstyle{remark}

\usepackage[textsize=tiny]{todonotes}

\begin{document}

\twocolumn[
\icmltitle{\model: Retrieval-Augmented Black-Box Language Models}



\icmlsetsymbol{equal}{*}
\icmlsetsymbol{thanks}{*}

\begin{icmlauthorlist}
\icmlauthor{Weijia Shi,\!}{uw,thanks}
\icmlauthor{Sewon Min,\!}{uw}
\icmlauthor{Michihiro Yasunaga,\!}{su}
\icmlauthor{Minjoon Seo,\!}{kaist}
\icmlauthor{Rich James,\!}{fb}
\icmlauthor{Mike Lewis,\!}{fb}
\icmlauthor{Luke Zettlemoyer\!}{uw,fb}
\icmlauthor{Wen-tau Yih}{fb}
\end{icmlauthorlist}

\icmlaffiliation{su}{Stanford University}
\icmlaffiliation{fb}{Meta AI}
\icmlaffiliation{uw}{University of Washington}
\icmlaffiliation{kaist}{KAIST }

\icmlcorrespondingauthor{Weijia Shi}{swj0419@uw.edu}


\vskip 0.3in
]



\printAffiliationsAndNotice{}  

\begin{abstract}
We introduce \model, a retrieval-augmented language modeling framework that treats the language model (LM) as a black box and augments it with a tuneable retrieval model. Unlike prior retrieval-augmented LMs that train language models with special cross attention mechanisms to encode the retrieved text, \model simply prepends retrieved documents to the input for the frozen black-box LM. 
This simple design can be easily applied to any existing retrieval and language models. 
Furthermore, we show that the LM can be used to supervise the retrieval model, which can then find documents that help the LM make better predictions. 
Our experiments demonstrate that \model with the tuned retriever 
significantly improves the performance of GPT-3 (175B) on language modeling by 6.3\%, as well as the performance of Codex on five-shot MMLU by 5.1\%. 


\end{abstract}

\section{Introduction}



 


Large language models (LLMs) such as GPT-3~\cite{NEURIPS2020_1457c0d6} and Codex~\cite{DBLP:journals/corr/abs-2107-03374}, 
have demonstrated impressive performance on a wide range of language tasks.
These models are typically trained on very large datasets and store a substantial amount of world or domain knowledge implicitly in their parameters. However, they are also prone to hallucination and cannot represent the full long tail of knowledge from the training corpus. Retrieval-augmented language models~\cite{Khandelwal2020Generalization, borgeaud2022improving, 
izacard2022few, yasunaga2022retrieval}, in contrast, can retrieve knowledge from an external datastore when needed, potentially reducing hallucination and increasing coverage. 
Previous approaches of retrieval-augmented language models require access to the internal LM representations (e.g., to train the model~\cite{borgeaud2022improving, izacard2022few} or to index the datastore~\cite{Khandelwal2020Generalization}), and are thus difficult to be applied to very large LMs. 
In addition, many best-in-class LLMs can only be accessed through APIs. Internal representations of such models are not exposed and fine-tuning is not supported.

\begin{figure}[]
\centering 
\includegraphics[scale=0.455]{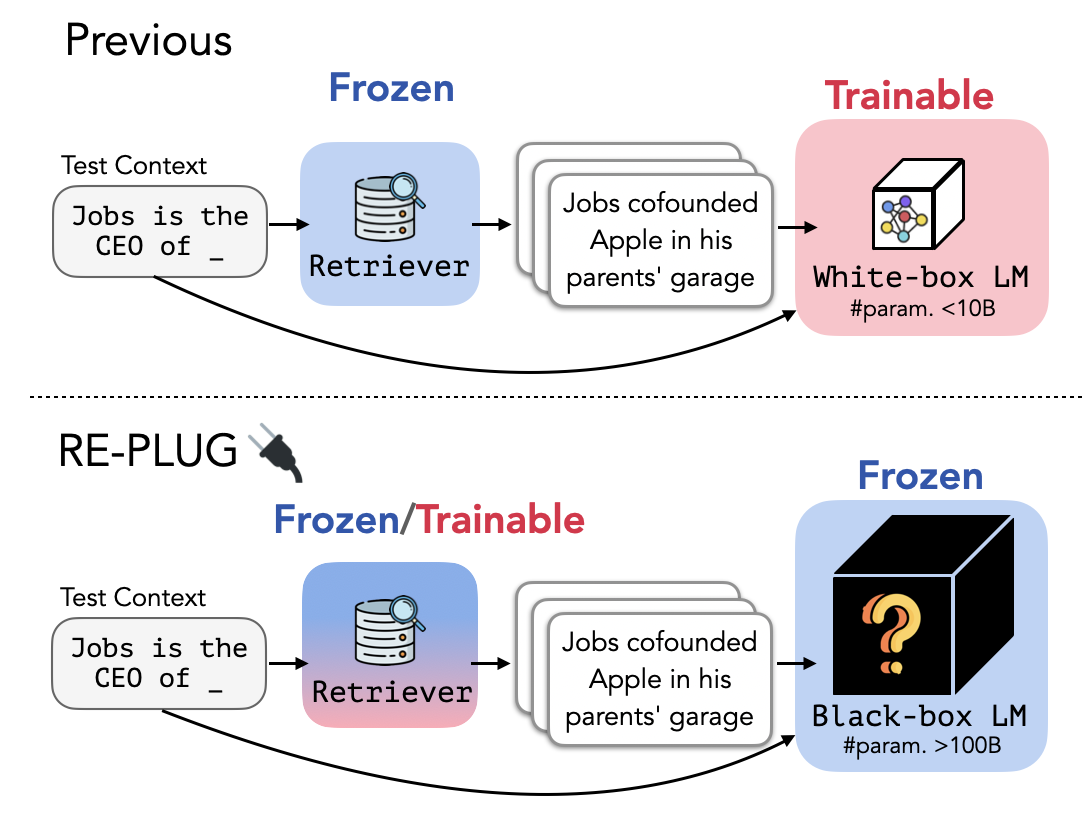}
\caption{
{Different from previous retrieval-augmented approaches~\cite{borgeaud2022improving} that enhance a language model with retrieval by updating the LM's parameters, \model treats the language model as a black box and augments it with a frozen or tunable retriever.} 
This black-box assumption makes \model applicable to large LMs (i.e., >100B parameters), which are often served via APIs.
} \label{fig:intro}
\end{figure}

In this work, we introduce \model (\textbf{Re}trieve and \textbf{Plug}), a new retrieval-augmented LM framework where the language model is viewed as a black box and the retrieval component is added as a tuneable plug-and-play module.
 Given an input context, \model first retrieves relevant documents from an external corpus using an \textit{off-the-shelf} retrieval model. The retrieved documents are prepended to the input context and fed into the black-box LM to make the final prediction. Because the LM context length limits the number of documents that can be prepended, we also introduce a new ensemble scheme that encodes the retrieved documents in parallel with the same black-box LM, allowing us to easily trade compute for accuracy. 
 As shown in \autoref{fig:intro}, \model is extremely flexible and can be used with any existing black-box LM and retrieval model.


 We also introduce \tunemodel (\model with \textbf{L}M-\textbf{S}upervised \textbf{R}etrieval), a training scheme that can further improve the initial retrieval model in \model with supervision signals from a black-box language model. 
 The key idea is to adapt the retriever to the LM, which is in contrast to prior work~\cite{borgeaud2022improving} that adapts language models to the retriever. 
We use a training objective which prefers retrieving documents that improve language model perplexity, while treating the LM as a frozen, black-box scoring function. 

Our experiments show that \model can improve the performance of diverse black-box LMs on both language modeling 
and downstream tasks, including MMLU~\cite{hendrycks2021measuring} and open-domain QA~\cite{kwiatkowski-etal-2019-natural, joshi-etal-2017-triviaqa}. 
For instance, \model can improve Codex (175B) performance on MMLU by 4.5\%, achieving comparable results to the 540B, instruction-finetuned Flan-PaLM.
Furthermore, tuning the retriever with our training scheme (i.e., \tunemodel) leads to additional improvements, including up to 6.3\% increase in GPT-3 175B language modeling.
To the best of our knowledge, our work is the first to show the benefits of retrieval to large LMs (>100B model parameters), for both reducing LM perplexity and and improving in-context learning performance. We summarize our contributions as follows:
\begin{itemize}
\item We introduce \model (\S \ref{section:inference}), the first retrieval-augmented language modeling framework for enhancing large black-box language models with retrieval. 
\item We propose a training scheme (\S \ref{section:tunemodel}) to further adapt an off-the-shelf retrieval model to the LM, using the language modeling scores as supervision signals, resulting in improved retrieval quality.

\item Evaluations on language modeling (\S \ref{section:experiments}), open-domain QA and MMLU demonstrate that \model can improve the performance of various language models such as GPT, OPT and BLOOM, including very large models with up to 175B parameters. 

\end{itemize}

\section{Background and Related Work}
\begin{figure*}
    \centering
    \includegraphics[scale=0.55]{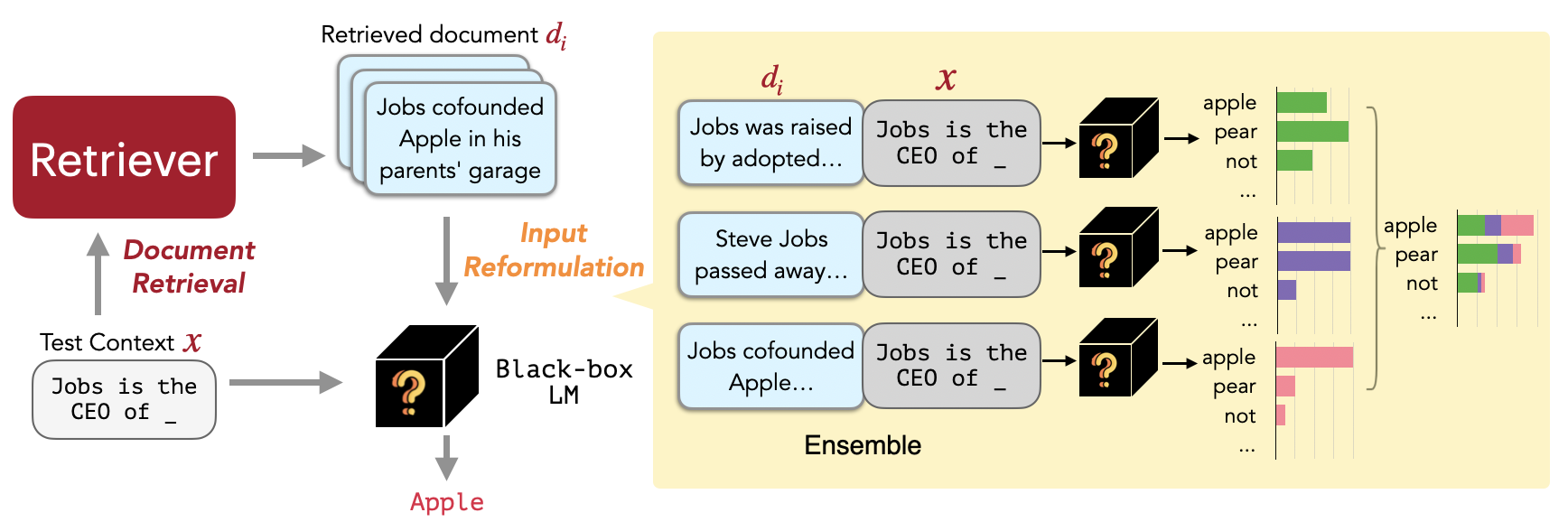}
    \caption{\textbf{\model at inference} (\S \ref{section:inference}). Given an input context, \model first retrieves a small set of relevant documents from an external corpus using a retriever (\S \ref{section:retrieve} \textit{Document Retrieval}). Then it prepends each document separately to the input context and ensembles output probabilities from different passes (\S \ref{section:fusion}  \textit{Input Reformulation}).
    }
    \label{fig:inference}
\end{figure*}



\paragraph{Black-box Language Models}

Large language models (i.e., >100B), such as GPT-3~\cite{NEURIPS2020_1457c0d6}, Codex~\cite{DBLP:journals/corr/abs-2107-03374}, and Yuan 1.0~\cite{wu2021yuan}, are not open-sourced due to commercial considerations and are only available as black-box APIs, through which users can send queries and receive responses. 
On the other hand, even open sourced language models such as OPT-175B~\cite{zhang2022democratizing} and BLOOM-176B~\cite{scao2022bloom} require significant computational resources to run and finetune locally. 
For example, finetuning BLOOM-176B requires 72 A100 GPUs (80GB memory, \$15k each~\cite{8bit}),
making them inaccessible to researchers and developers with limited resources.
Traditionally, retrieval-augmented model frameworks~\cite{Khandelwal2020Generalization, borgeaud2022improving, yu-2022-retrieval, izacard2022few, goyal2022retrieval} have focused on the white-box setting, where language models are fine-tuned to incorporate retrieved documents. However, the increasing scale and black-box nature of large language models makes this approach infeasible. To address the challenges posed by large language models, we investigate retrieval-augmentation in the \textbf{black-box setting}, where users only have access to the model predictions and cannot access or modify its parameters.

\paragraph{Retrieval-augmented Models} 
Augmenting language models with relevant information retrieved from various knowledge stores has shown to be effective in improving performance on various NLP tasks, including language modeling~\cite{min2022nonparametric, borgeaud2022improving, Khandelwal2020Generalization} and open-domain question answering~\cite{rag2020, izacard2022few, hu2022promptcap}. 
Specifically, using the input as query, (1) a retriever first retrieves a set of documents (i.e., sequences of tokens) from a corpus and then (2) a language model incorporates the retrieved documents as additional information to make a final prediction. 
This style of retrieval can be added to both \emph{encoder-decoder}~\cite{yu-2022-retrieval, izacard2022few} 
and \textit{decoder-only} models~\cite{Khandelwal2020Generalization, borgeaud2022improving, knnprompt, rubin2022learning}. For example, Atlas~\cite{izacard2022few} finetunes an \emph{encoder-decoder} model jointly with the retriever by modeling documents as latent variables, while RETRO~\cite{borgeaud2022improving} changes the \textit{decoder-only} architecture to incorporate retrieved texts and pretrains the language model from scratch. 
Both methods require updating the model parameters through gradient descent, which cannot be applied to black-box LMs. 
Another line of retrieval-augmented LMs such as kNN-LM~\cite{Khandelwal2020Generalization, zhong2022training} retrieves a set of tokens and interpolates between the LM's next token distribution and kNN distributions computed from the retrieved tokens at inference. 
Although kNN-LM does not require additional training, it requires access to internal LM representations to compute the kNN distribution, which are not always available for large LMs such as GPT-3. In this work, we investigate ways to improve large black-box language models with retrieval.
While concurrent work~\cite{mallen2022not, chenglei, yu2022generate, khattab2022demonstrate} has demonstrated that using a frozen retriever can improve GPT-3 performance on open-domain question answering, we approach the problem in a more general setting, including language modeling and understanding tasks.
We also propose an ensemble method to incorporate more documents and a training scheme to further adapt the retriever to large LMs.

\section{\model} \label{section:inference}
We introduce \model (\textbf{Re}trieve and \textbf{Plug}), a new retrieval-augmented LM paradigm where the language model is treated as black box and the retrieval component is added as a potentially tuneable module.

As shown in Figure~\ref{fig:inference}, given an input context, \model first retrieves a small set of relevant documents from an external corpus
using a retriever (\S \ref{section:retrieve}). Then
we pass the concatenation of each retrieved document with the input context through the LM in parallel, and ensemble the predicted probabilities (\S \ref{section:fusion}).


\subsection{Document Retrieval} 
\label{section:retrieve}
Given an input context $x$, the retriever aims to retrieve a small set of documents from a corpus $\mathcal{D}=\{d_1...d_m\}$ that are relevant to $x$. 
Following prior work~\cite{qu-etal-2021-rocketqa, izacard2021leveraging, gtr21}, 
we use a dense retriever based on the dual encoder architecture, where an encoder is used to encode both the input context $x$ and the document $d$.
Specifically, the encoder maps each document $d \in D$ to an embedding $\textbf{E}(d)$ by taking the mean pooling of the last hidden representation over the tokens in $d$. 
At query time, the same encoder is applied to the input context $x$ to obtain a query embedding $\textbf{E}(x)$. 
The similarity between the query embedding and the document embedding is computed by their cosine similarity:
\begin{align}
s(d, x) = \cos (\textbf{E}(d), \textbf{E}(x))  \label{eq:sim}
\end{align}
The top-$k$ documents that have the highest similarity scores when compared with the input $x$ are retrieved in this step. For efficient retrieval, we precompute the embedding of each document $d \in D$ and construct FAISS index~\cite{johnson2019billion} over these embeddings.

\subsection{Input Reformulation} \label{section:fusion}
The retrieved top-$k$ documents provide rich information about the original input context $x$ and can potentially help the LM to make a better prediction.
One simple way to incorporate the retrieved documents as part of the input to the LM is to prepend $x$ with all $k$ documents.
However, this simple scheme is fundamentally restricted by the number of documents (i.e., $k$) we can include, given the language model's context window size. 
To address this limitation, we adopt an ensemble strategy 
described as follows. 
Assume $\mathcal{D}' \subset \mathcal{D}$ consists of $k$ most relevant documents to $x$, according to the scoring function in Eq.~\eqref{eq:sim}. 
We prepend each document $d \in \mathcal{D}'$ to $x$, pass this concatenation to the LM separately, and then ensemble output probabilities from all $k$ passes. 
Formally, given the input context $x$ and its top-$k$ relevant documents $\mathcal{D}'$, the output probability of the next token $y$ is computed as a weighted average ensemble:
\begin{align*}
p(y \mid x, \mathcal{D}') = \sum_{d \in \mathcal{D}'} p(y \mid d \circ x) \cdot \lambda(d,x),
\end{align*}
where $\circ$ denotes the concatenation of two sequences and the weight $\lambda(d,x)$ is based on the similarity score between the document $d$ and the input context $x$:
\begin{align*}
\lambda(d,x) = \frac{e^{s(d,x)}}{\sum_{d \in \mathcal{D}'}e^{s(d,x)}}
\end{align*}
Although our ensemble method requires running the LM $k$ times, the cross attention is performed between each retrieved document and the input context. 
Therefore, compared with the method of prepending all the retrieved documents, our ensemble methods do not incur additional computational cost overhead. 


\begin{figure*}
    \centering
    \includegraphics[scale=0.53]{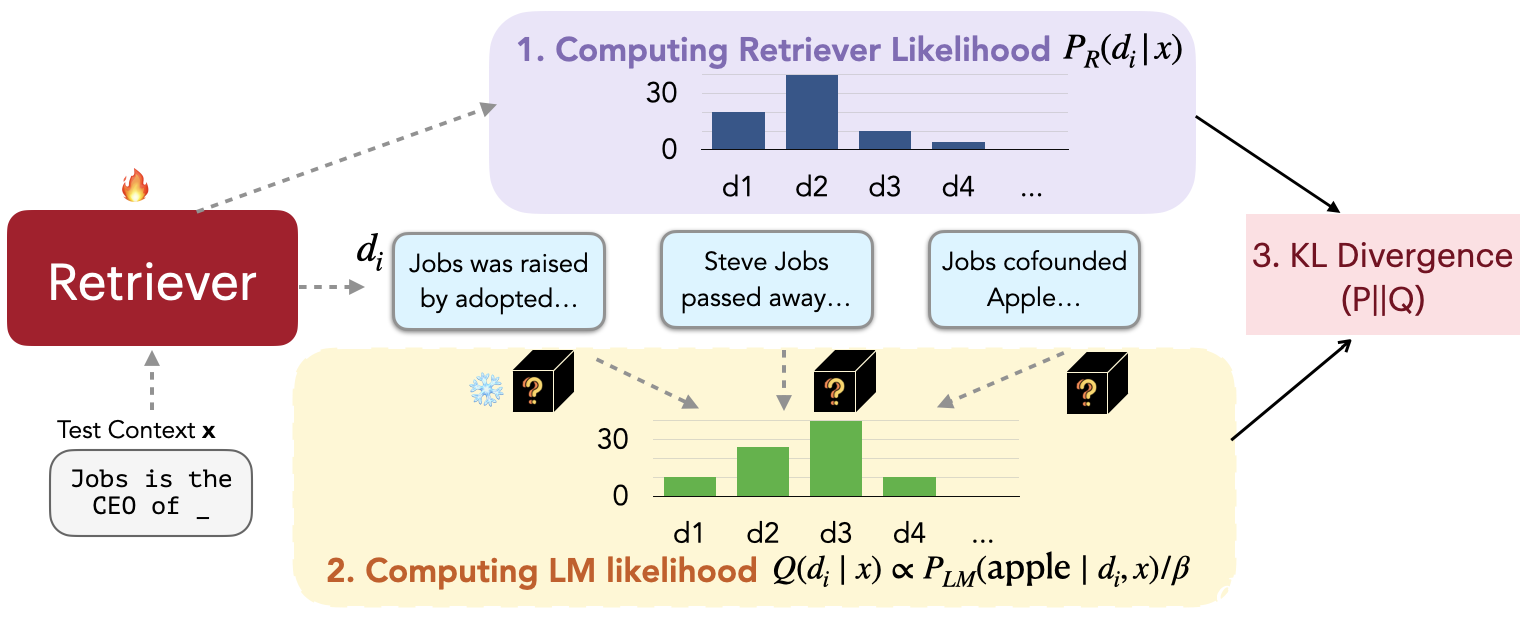}
    \caption{\textbf{\tunemodel training process (\S\ref{section:tunemodel}).} The retriever is trained using the output of a frozen language model as supervision signals.}
    \label{fig:knn_prompt}
\end{figure*}


\section{\tunemodel: Training the Dense Retriever} \label{section:tunemodel}

Instead of relying only on existing neural dense retrieval models  ~\cite{karpukhin-etal-2020-dense, izacard2022unsupervised, su2022one}, 
we further propose \tunemodel (\model with LM-Supervised Retrieval), which \emph{adapts} the retriever in \model by using the LM itself to provide supervision about which documents should be retrieved. 

Inspired by \citet{sachan2022questions}, our approach can be seen as adjusting the probabilities of the retrieved documents to match the probabilities of the output sequence perplexities of the language model. In other words, we would like the retriever to find documents that result in lower perplexity scores.
As shown in Figure~\ref{fig:knn_prompt}, our training algorithm consists of the four steps: (1) retrieving documents and computing the retrieval likelihood (\S \ref{section:retrieval likelihood}), (2) scoring the retrieved documents by the language model (\S \ref{section:LM score}), (3) updating the retrieval model parameters by minimizing the KL divergence between the retrieval likelihood and the LM's score distribution (\S \ref{section:loss}), and (4) asynchronous update of the datastore index (\S \ref{section:update}). 

\subsection{Computing Retrieval Likelihood} 
\label{section:retrieval likelihood}
We retrieve $k$ documents $\mathcal{D}' \subset \mathcal{D}$ with the highest similarity scores from a corpus $\mathcal{D}$ given an input context $x$, as described in \S \ref{section:retrieve}. We then compute the retrieval likelihood of each retrieved document $d$:
\begin{align*}
P_{R}(d \mid x) = \frac{e^{s(d,x)/\gamma}}{\sum_{d \in \mathcal{D}'}e^{s(d,x)/\gamma}}
\end{align*}
where $\gamma$ is a hyperparameter that controls the temerature of the softmax. Ideally, the retrieval likelihood is computed by marginalizing over all the documents in the corpus $\mathcal{D}$, which is intractable in practice. Therefore, we approximate the retrieval likelihood by only marginalizing over the retrieved documents $\mathcal{D}'$.

\subsection{Computing LM likelihood} \label{section:LM score}
We use the LM as a scoring function to measure how much each document could improve the LM perplexity. Specifically, we first compute $P_{LM}(y \mid d, x)$, the LM probability of the ground truth output $y$ given the input context $x$ and a document $d$. The higher the probability, the better the document $d_i$ is at improving the LM's perplexity. We then compute the LM likelihood of each document $d$ as follows:
\begin{align*}
Q(d \mid x, y) = \frac{e^{P_{LM}(y \mid d, x)/\beta}}{\sum_{d \in \mathcal{D}'}e^{P_{LM}( y \mid d, x)/\beta}}
\end{align*}
where $\beta$ is another hyperparameter.

\subsection{Loss Function}\label{section:loss}
Given the input context $x$ and the corresponding ground truth continuation $y$, we compute the retrieval likelihood and the language model likelihood. The dense retriever is trained by minimizing the KL divergence between these two distributions:
\begin{align*}
\mathcal{L} = \frac{1}{|\mathcal{B}|} \sum_{x \in \mathcal{B}}{ KL \Big( P_{R}(d \mid x) \parallel Q_{\text{LM}}(d \mid x, y) \Big) },
\end{align*}
where $\mathcal{B}$ is a set of input contexts.
When minimizing the loss, we can only update the retrieval model parameters. The LM parameters are fixed due to our black-box assumption.

\subsection{Asynchronous Update of the Datastore Index} \label{section:update}
Because the parameters in the retriever are updated during the training process, the previously computed document embeddings are no longer up to date. Therefore, following~\citet{guu2020retrieval}, we recompute the document embeddings and rebuild the efficient search index using the new embeddings every $T$ training steps. Then we use the new document embeddings and index for retrieval, and repeat the training procedure.

\section{Training Setup} 
\label{section:training details}
In this section, we describe the details of our training procedure. We first describe the model setting in \model (\S \ref{subsection:model}) and then describe the procedure for training the retriever in \tunemodel (\S \ref{subsection:tunemodel}). 


\subsection{\model} \label{subsection:model}
In theory, any type of retriever, either dense~\cite{karpukhin2020dense, gtr21} or sparse~\cite{robertson2009probabilistic}, could be used for \model.
Following prior work~\cite{izacard2022few}, 
we use the Contriever~\cite{izacard2022unsupervised} as the retrieval model for \model, as it has demonstrated strong performance. 

\vspace{-1mm}
\subsection{\tunemodel} \label{subsection:tunemodel}
For \tunemodel, we initialize the retriever with the Contriever model~\cite{izacard2022unsupervised}. We use GPT-3 Curie~\cite{gpt3} as the supervision LM to compute the LM likelihood. 

\vspace{-2mm}
\paragraph{Training data}
We use 800K sequences of 256 tokens each, sampled from the Pile training data~\cite{pile}, as our training queries. Each query is split into two parts: the first 128 tokens are used as the input context $x$, and the last 128 tokens are used as the ground truth continuation $y$. For the external corpus $D$, we sample 36M documents of 128 tokens from the Pile training data. To avoid trivial retrieval, we ensure that the external corpus documents do not overlap with the documents from which the training queries are sampled.


\vspace{-2mm}
\paragraph{Training details}
To make the training process more efficient, we pre-compute the document embeddings of the external corpus $D$ and create a FAISS index~\cite{johnson2019billion} for fast similarity search. Given a query $x$, we retrieve the top 20 documents from the FAISS index and compute the retrieval likelihood and the LM likelihood with a temperature of 0.1. We train the retriever using the Adam optimizer~\cite{kingma2015adam} with a learning rate of 2e-5, a batch size of 64, and a warmup ratio of 0.1. We re-compute the document embeddings every 3k steps and fine-tune the retriever for a total of 25k steps.




\begin{table*}[h]
\small
\centering
\begin{tabular}{llcc|cc|cc}
\toprule
Model & & \# Parameters & Original & + \model\ & Gain \% & + \tunemodel\ & Gain \% \\
\midrule
GPT-2 & Small & 117M & 1.33 & 1.26 & 5.3 & 1.21 & 9.0 \\
& Medium & 345M & 1.20 & 1.14 & 5.0 & 1.11 & 7.5 \\
& Large & 774M & 1.19 & 1.15 & 3.4 & 1.09 & 8.4 \\
& XL & 1.5B & 1.16 & 1.09 & 6.0 & 1.07 & 7.8 \\
\midrule
GPT-3 & Ada & 350M & 1.05 & 0.98 & 6.7 & 0.96 & 8.6 \\
(black-box) & Babbage & 1.3B & 0.95 & 0.90 & 5.3 & 0.88 & 7.4 \\
& Curie & 6.7B & 0.88 & 0.85 & 3.4 & 0.82 & 6.8 \\
& Davinci & 175B & 0.80 & 0.77 & 3.8 & 0.75 & 6.3 \\
\bottomrule
\end{tabular}
\caption{\textbf{Both \model and \tunemodel consistently enhanced the performance of different language models.} 
 Bits per byte (BPB) of the Pile using GPT-3 and GPT-2 family models (Original) and their retrieval-augmented versions (+\model and +\tunemodel. The gain \% shows the relative improvement of our models compared to the original language model. 
} 
\label{tab:main}
\end{table*}

\section{Experiments}
\label{section:experiments}
We perform evaluations on both language modeling (\S \ref{section:LM}) and downstream tasks such as MMLU (\S \ref{section:MMLU}) and open-domain QA (\S \ref{section:QA}). In all settings, \model\~ improve the performance of various black-box language models, showing the effectiveness and  generality of our approach. 

\subsection{Language Modeling} \label{section:LM}


\paragraph{Datasets}
The Pile~\cite{pile} is a language modeling benchmark that consists of text sources from diverse domains such as web pages, code and academic papers. Following prior work, we report bits per UTF-8 encoded byte (BPB) as the metric on each subset domain. 

\paragraph{Baselines}
We consider GPT-3 and GPT-2 family language model as the baselines. The four models from GPT-3 (Davinci, Curie, Baddage and Ada) are black-box models that are only accessible through API 

\paragraph{Our model}
We add \model and \tunemodel to the baselines. We randomly subsampled Pile training data (367M documents of 128 tokens) and use them as the retrieval corpus for all models. As the Pile dataset has made efforts to deduplicate documents across train, validation and test splits~\cite{pile}, we did not do additional filtering. For both \model and \tunemodel, we use a length of 128-token context to do retrieval and adopt the ensemble method (Section ~\ref{section:fusion}) to incorporate top 10 retrieved documents during inference.


\paragraph{Results}
Table~\ref{tab:main} reports the results of the original baselines, baselines augmented with the \model, and baselines augmented with the \tunemodel. We observe that both \model and \tunemodel significantly outperform the baselines. This demonstrates that simply adding a retrieval module to a frozen language model (i.e., the black-box setting) is effective at improving the performance of different sized language models on language modeling tasks. Furthermore, \tunemodel consistently performs better than \model by a large margin. Specifically, \tunemodel results in 7.7\% improvement over baselines compared to 4.7\% improvement of \model averaged over the 8 models. This indicates that further adapting the retriever to the target LM is beneficial.

\subsection{MMLU} \label{section:MMLU}

\begin{table*}[h]
\small
 \addtolength{\tabcolsep}{-1pt} 
\centering
\begin{tabular}{llcccc|c}
\toprule
Model & \# Parameters & Humanities & Social. & STEM & Other & All \\
\midrule
Codex & 175B & 74.2 & 76.9 & 57.8 & 70.1 & 68.3 \\
PaLM & 540B & 77.0 & 81.0 & 55.6 & 69.6 & 69.3 \\
Flan-PaLM & 540B & - & - & - & - & 72.2 \\
\midrule
Atlas & 11B & 46.1 & 54.6 & 38.8 & 52.8 & 47.9 \\
\midrule
Codex + \model\ & 175B & 76.0 & 79.7 & 58.8 & 72.1 & 71.4 \\
Codex + \tunemodel\  & 175B & 76.5 & 79.9 & 58.9 & 73.2 & 71.8 \\
\bottomrule
\end{tabular}
\caption{\textbf{\model and \tunemodel improves Codex by 4.5\% and 5.1\% respectively.} Performance on MMLU broken down into 4 categories. The last column averages the performance over these categories. All models are evaluated based on 5-shot in-context learning with direct prompting. 
}
\label{tab:mmlu}
\end{table*}

\paragraph{Datasets}
 Massive Multi-task Language Understanding (MMLU~\cite{hendrycks2021measuring}) is a multiple choice QA dataset that covers exam questions from 57 tasks including mathematics, computer science, law, US history and etc. The 57 tasks are grouped into 4 categories: humanities, STEM, social sciences and other. Following \citet{chung2022scaling}, we evaluate \model in the 5-shot in-context learning setting.
 

\paragraph{Baselines}
We consider two groups of strong previous models as baselines for comparisons. The first group of baselines is the 
state-of-the-art LLMs including Codex\footnote{Code-Davinci-002}~\cite{codex}, PaLM~\cite{palm}, and Flan-PaLM~\cite{flan-palm}. According to \citet{flan-palm}, these three models rank top-3 in the leaderboard of MMLU. The second group of baselines consists of retrieval-augmented language models. We only include Atlas~\cite{izacard2022few} in this group, as no other retrieval-augmented LMs have been evaluated on the MMLU dataset. Atlas trains both the retriever and the language model, which we consider a white-box retrieval LM setting.

\paragraph{Our model}
We add \model and \tunemodel only to Codex because other models such as PaLM and Flan-PaLM are not accessible to the public. We use the test question as the query to retrieve 10 relevant documents from Wikipedia (2018, December) and prepend each retrieved document to the test question, resulting in 10 separate inputs.
These inputs are then separately fed into the language models, and the output probabilities are ensemble together. 

\paragraph{Results}
Table \ref{tab:mmlu} presents the results from the baselines, \model, and \tunemodel on the MMLU dataset. 
We observe that both the \model and \tunemodel improve the original Codex model by 4.5\% and 5.1\%, respectively. In addition, \tunemodel largely outperforms the previous retrieval-augmented language model, Atlas, demonstrating the effectiveness of our black-box retrieval language model setting.
Although our models slightly underperform Flan-PaLM, this is still a strong result because Flan-PaLM has three times more parameters. We would expect that the \tunemodel could further improve Flan-PaLM, if we had access to the model.

Another interesting observation is that the \tunemodel outperforms the original model by 1.9\% even in the STEM category. This suggests that retrieval may improve a language model's problem-solving abilities.


\subsection{Open Domain QA} \label{section:QA}
Lastly, we conduct evaluation on two open-domain QA datasets: Natural Questions (NQ)~\cite{kwiatkowski-etal-2019-natural} and TriviaQA~\cite{joshi-etal-2017-triviaqa}. 

\begin{table}[h]
\small
 \addtolength{\tabcolsep}{-1pt} 
\centering
\begin{tabular}{lcccc}
\toprule
  & \multicolumn{2}{c}{NQ} & \multicolumn{2}{c}{TQA} \\
\cmidrule(lr){2-3}
\cmidrule(lr){4-5}
Model  & Few-shot & Full & Few-shot & Full \\
\midrule
Chinchilla  & 35.5 & - & 64.6 & - \\
PaLM  & 39.6 & - & - & - \\
Codex  & 40.6 & - & 73.6 & - \\
\midrule
RETRO$^\dag$  & - & 45.5 & - & - \\
R2-D2$^\dag$ & - & 55.9 & - & 69.9 \\
Atlas$^\dag$ & 42.4 & \textbf{60.4} & 74.5 & \textbf{79.8} \\
Codex + Contriever$_{cc}$\tablefootnote{\citet{si2022prompting} augment Codex with concatenation of 10 documents retrieved by contriever.} & 44.2 & - & 76.0 & - \\
\midrule
Codex + \model\  & 44.7 & - & 76.8 & - \\
Codex + \tunemodel\ & \textbf{45.5} & - & \textbf{77.3} & - \\
\bottomrule
\end{tabular}
\caption{Performance on NQ and TQA. We report results for both few-shot (64 shots for Chinchilla, PaLM, and Atlas; 16 shots for Codex-based models) and full training data settings. \tunemodel improves Codex by 12.0\% on NQ and 5.0\% on TQA, making it the best-performing model in the few-shot setting. Note that models with \dag~are finetuned using training examples, while other models use in-context learning. 
}
\label{tab:qa}
\end{table}

\paragraph{Datasets}
NQ and TriviaQA are two open-domain QA datasets consisting of questions, answers collected from Wikipedia and the Web. Following prior work~\cite{izacard-grave-2021-leveraging, si2022prompting}, we report results for the filtered set of TriviaQA. For evaluation, we consider the few-shot setting where the model is only given a few training examples and full data where the model is given all the training examples.

\paragraph{Baselines}
We compare our model with several state-of-the-art baselines, both in a few-shot setting and with full training data. The first group of models consists of powerful large language models, including Chinchilla~\cite{chinchilla}, PaLM~\cite{palm}, and Codex. These models are all evaluated using in-context learning under the few-shot setting, with Chinchilla and PaLM evaluated using 64 shots, and Codex using 16 shots. The second group of models for comparison includes retrieval-augmented language models such as RETRO~\cite{borgeaud2021improving}, R2-D2~\cite{fajcik-etal-2021-r2-d2}, and Atlas~\cite{izacard2022few}. All of these retrieval-augmented models are finetuned on the training data, either in a few-shot setting or with full training data. Specifically, Atlas is finetuned on 64 examples in the few-shot setting.

\paragraph{Our model}
We add \model and \tunemodel to Codex with Wikipedia (2018, December) as the retrieval corpus to evaluate the model in a 16-shot in context learning. Similar to the setting in language modeling and MMLU, we incorporate top-10 retrieved documents using our proposed ensemble method. 

\paragraph{Results}


As shown in Table~\ref{tab:qa}, \tunemodel significantly improves the performance of the original Codex by 12.0\% on NQ and 5.0\% on TQA. It outperforms the previous best model, Atlas, which was fine-tuned with 64 training examples, achieving a new state-of-the-art in the few-shot setting. However, this result still lags behind the performance of retrieval-augmented language models fine-tuned on the full training data. This is likely due to the presence of near-duplicate test questions in the training set (e.g., \citet{lewis2021question} found that 32.5\% of test questions overlap with the training sets in NQ). 






\section{Analysis}

\begin{figure}[]
\centering 
\includegraphics[scale=0.34]{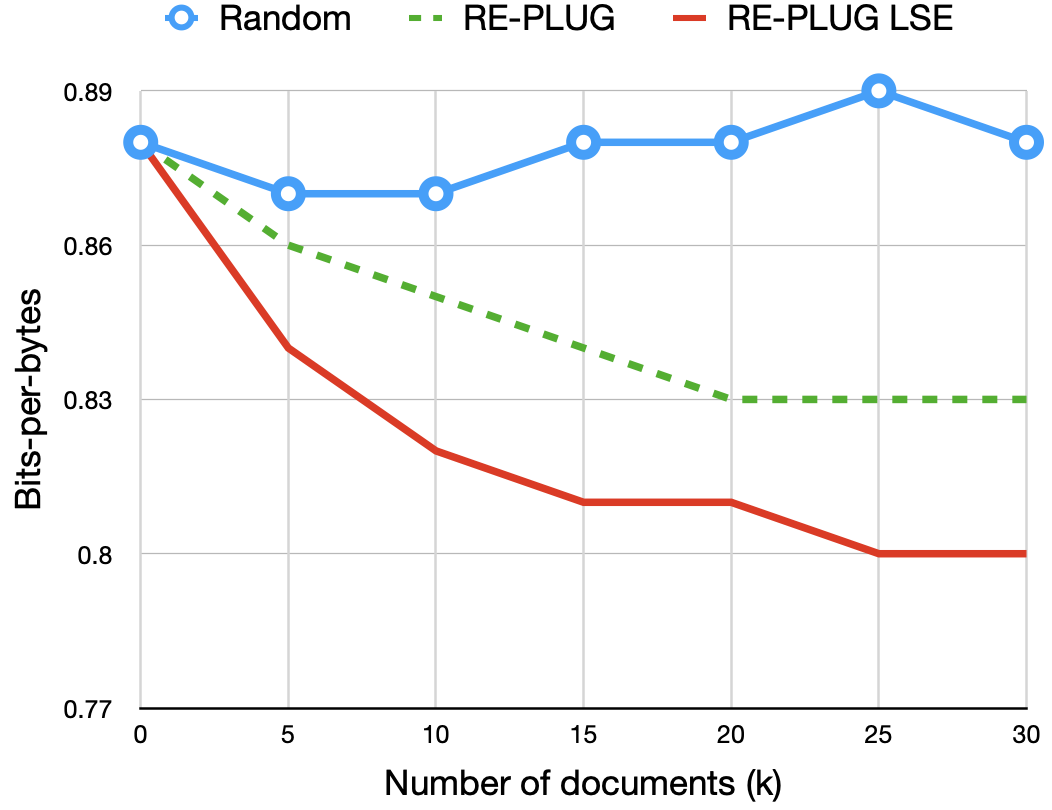}
\caption{
\textbf{Ensembling random documents does not result in improved performance.} BPB of Curie augmented with different methods (random, \model and \tunemodel) when varying the number of documents (i.e.; number of ensemble times.)}  
\label{fig:ensemble}
\end{figure}


\subsection{\model performance gain does not simply come from the ensembling effect
}

The core of our method design is the use of an ensemble method that combines output probabilities of different passes, in which each retrieved document is prepended separately to the input and fed into a language model. To study whether the gains come solely from the ensemble method, we compare our method to ensembling random documents. For this, we randomly sample several documents, concatenated each random document with the input, and ensemble the outputs of different runs (referred to as "random").
As shown in \autoref{fig:ensemble}, we evaluated the performance of GPT-3 Curie on Pile when augmented with random documents, documents retrieved by \model, and documents retrieved by \tunemodel. We observed that ensembling random documents leads to worse performance, indicating that the performance gains of \model do not solely come from the ensembling effect. Instead, ensembling the \textbf{relevant} documents is crucial for the success of \model. Additionally, as more documents were ensembled, the performance of \model and \tunemodel improved monotonically. However, a small number of documents (e.g., 10) was sufficient to achieve large performance gains.





\begin{figure*}[ht]
    \centering
    \subfigure[]{\includegraphics[width=0.3\textwidth]{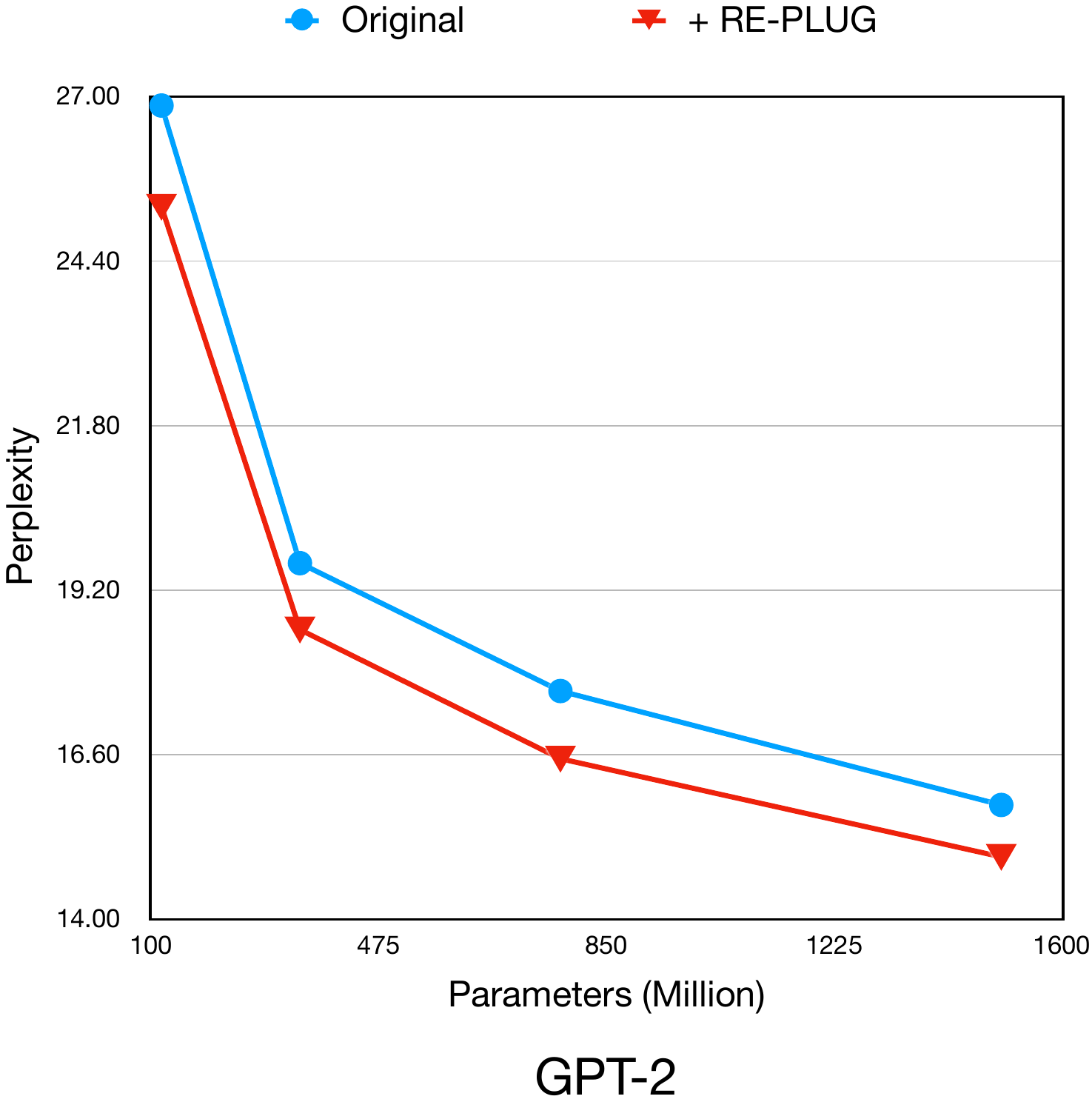}} 
    \subfigure[]{\includegraphics[width=0.3\textwidth]{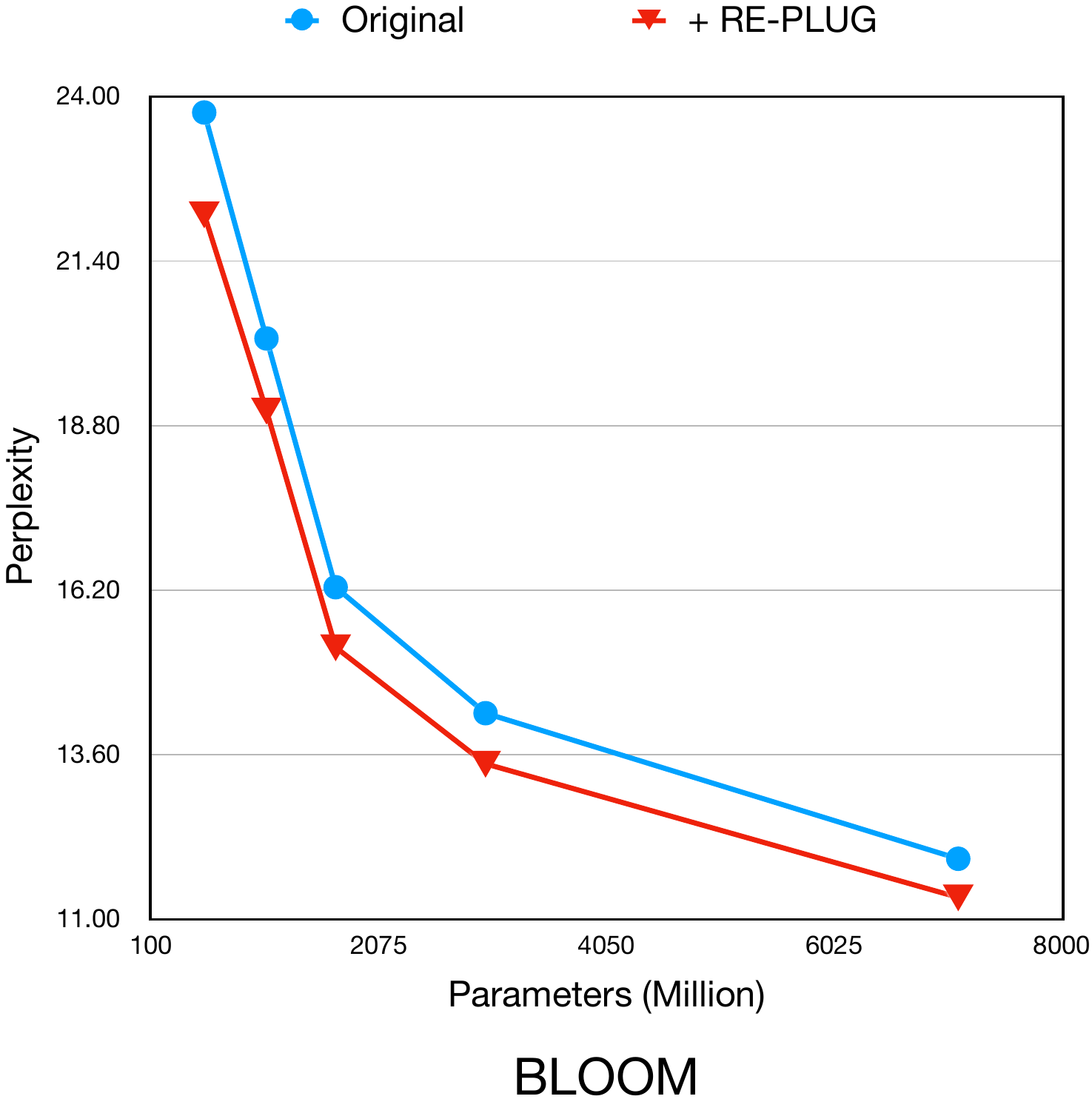}} 
    \subfigure[]{\includegraphics[width=0.3\textwidth]{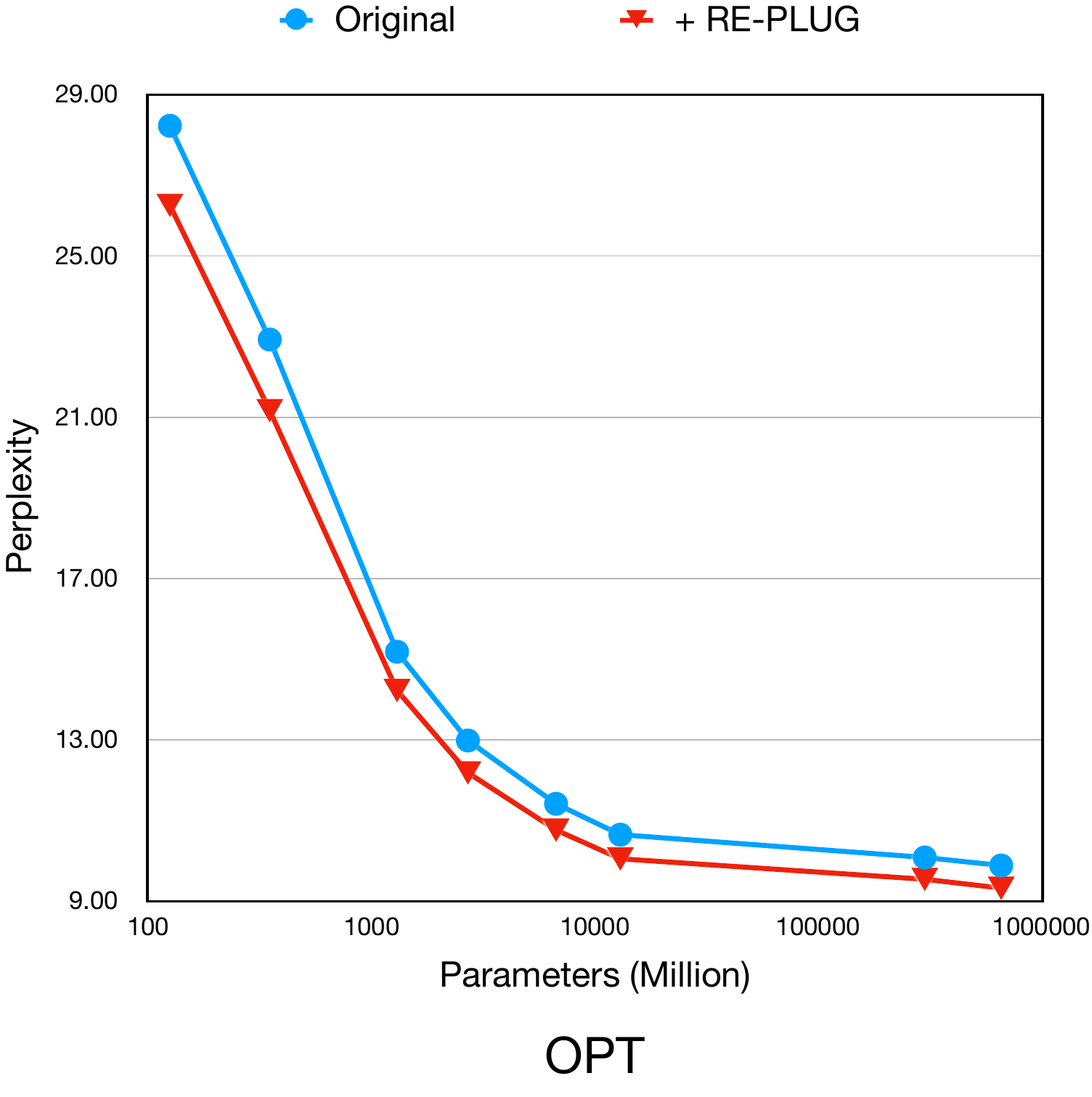}}
    \caption{
   \textbf{GPT-2, BLOOM and OPT models of varying sizes consistently benefit from \model.} 
 The x-axis indicates the size of the language model and the y-axis is its perplexity on Wikitext-103. 
    }
    \label{fig:scaling}
\end{figure*}

\subsection{\model is applicable to diverse language models}
Here we further study whether \model could enhance \textit{diverse} language model families that have been pre-trained using different data and methods.  Specifically, we focus on three groups of language models with varying sizes: GPT-2 (117M, 345M, 774M, 1.5B parameters)~\cite{NEURIPS2020_1457c0d6}, OPT (125M, 350M, 1.3B, 2.7B, 6.7B, 13B, 30B, 66B)~\cite{zhang2022opt} and BLOOM (560M, 1.1B, 1.7B, 3B and 7B)~\cite{scao2022bloom}. We evaluate each model on Wikitext-103~\cite{stephen2017pointer} test data and report its perplexity. For comparison, we augment each model with \model that adopts the ensemble method to incorporate top 10 retrieved documents. Following prior work~\cite{Khandelwal2020Generalization}, we use Wikitext-103 training data as the retrieval corpus. 

\autoref{fig:scaling} shows the performance of different-sized language models with and without \model. We observe that the performance gain brought by \model stays consistent with model size. For example, OPT with 125M parameters achieves 6.9\% perplexity improvement, while OPT with 66B parameters achieves 5.6\% perplexity improvement. Additionally, \model improves the perplexity of all the model families.
This indicates that \model is applicable to diverse language models with different sizes.


\subsection{Qualitative Analysis: rare entities benefit from retrieval}
To understand why the \model improves language modeling performance, we conducted manual analysis of examples in which the \model results in a decrease in perplexity. We find that \model is more helpful when texts contain rare entities.
\autoref{fig:ensemble} shows a test context and its continuation from the Wikitext-103 test set. For \model, we use the test context as a query to retrieve a relevant document from Wikitext-103 training data. We then compute the perplexity of the continuation using the original GPT-2 1.5B and its \model enhanced version. After incorporating the retrieved document, the perplexity of the continuation improves by 11\%. 
Among all tokens in the continuation, we found that \model is most helpful for the rare entity name "Li Bai".
This is likely because the original LM does not have sufficient information about this rare entity name. However, by incorporating the retrieved document, \model was able to match the name with the relevant information in the retrieved document, resulting in better performance.


\begin{figure}[h]
\centering 
\includegraphics[scale=0.29]{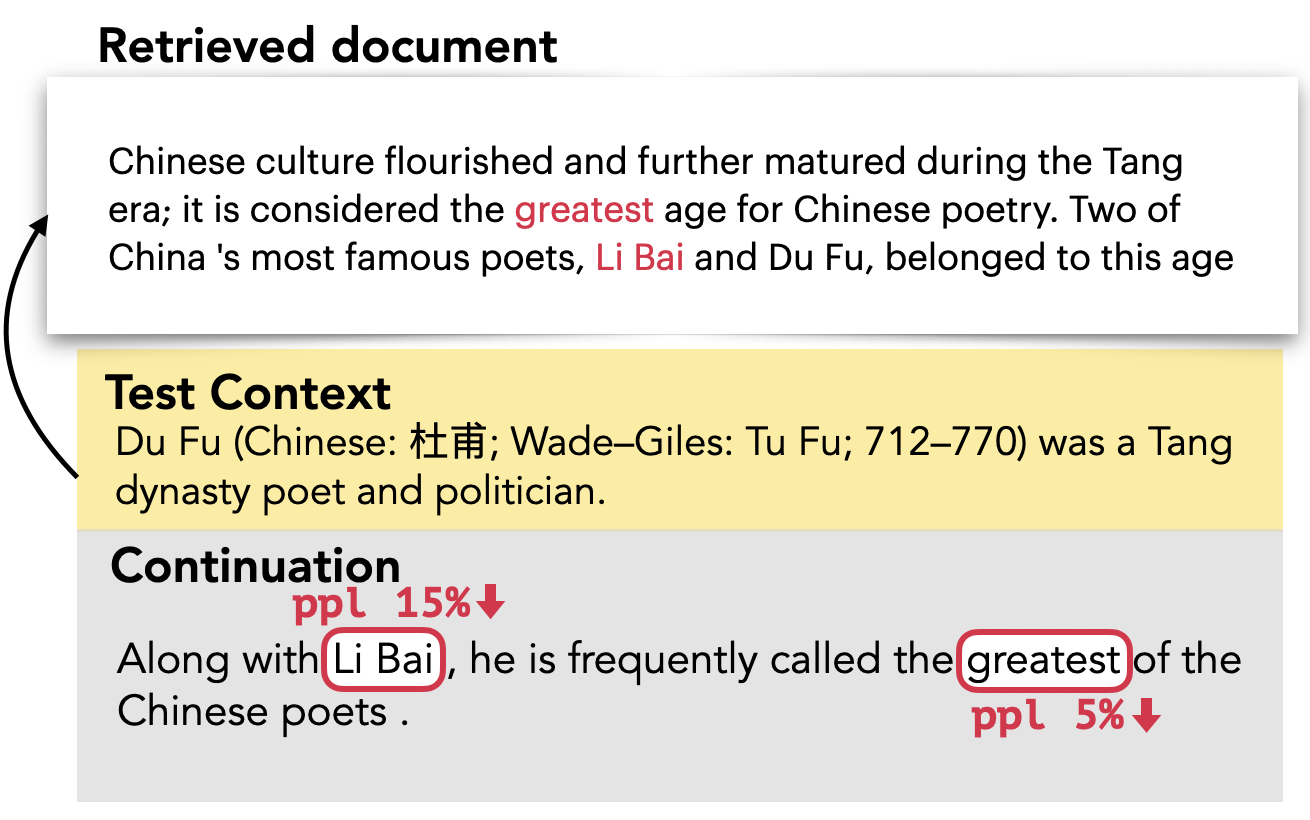}
\caption{\textbf{Rare entities benefit from
retrieval}. After incorporating the retrieved document during inference, the entity "\textit{Li Bai}" and the token "\textit{greatest}" in the continuation show the most improvement in perplexity (15\% for "\textit{Li Bai}" and 5\% for "\textit{greatest}"). Other tokens' perplexity changes are within 5\%.}
\label{fig:ensemble}
\end{figure}



\section{Conclusion}
We introduce \model, a retrieval-augmented language modeling paradigm that treats the language model as a black box and augments it with a tuneable retrieval model. Our evaluation shows that \model can be integrated with any existing language model to improve their performance on language modeling or downstream tasks. This work opens up new possibilities for integrating retrieval into large-scale black-box language models and demonstrates even the state-of-the-art large-scale LMs could benefit from retrieval. However, \model lacks interpretability as it is unclear when the model relies on retrieved knowledge or parametric knowledge. Future research could focus on developing more interpretable retrieval-augmented language models.



\bibliography{anthology,custom}
\bibliographystyle{icml2022}

\newpage

\appendix
\onecolumn

\begin{table*}
\small
\begin{tabularx}{\textwidth}{m{17cm}}
\toprule
\textbf{Knowledge}: Arctic Ocean. Although over half of Europe's original forests disappeared through the centuries of deforestation, Europe still has over one quarter of its land area as forest, such as the broadleaf and mixed forests, taiga of Scandinavia and Russia, mixed rainforests of the Caucasus and the Cork oak forests in the western Mediterranean. During recent times, deforestation has been slowed and many trees have been planted. However, in many cases monoculture plantations of conifers have replaced the original mixed natural forest, because these grow quicker. The plantations now cover vast areas of land, but offer poorer habitats for many European \\ 
\textbf{Question}: As of 2015, since 1990 forests have \xdash in Europe and have \xdash in Africa and the Americas. \\
A. "increased, increased"
B. "increased, decreased"
C. "decreased, increased"
D. "decreased, decreased" \\
\textbf{Answer}: B \\
\\
\textbf{Knowledge}: Over the past decades, the political outlook of Americans has become more progressive, with those below the age of thirty being considerably more liberal than the overall population. According to recent polls, 56\% of those age 18 to 29 favor gay marriage, 68\% state environmental protection to be as important as job creation, 52\% "think immigrants \'strengthen the country with their hard work and talents,\'" 62\% favor a "tax financed, government-administrated universal health care" program and 74\% "say \'people\'s will\' should have more influence on U.S. laws than the Bible, compared to 37\%, 49\%, 38\%, 47\% and 58\% among the \\ 
\textbf{Question}: As of 2019, about what percentage of Americans agree that the state is run for the benefit of all the people? \\
A. 31\%
B. 46\%
C. 61\%
D. 76\% \\
\textbf{Answer}: B \\
... \\
\textbf{Knowledge}: last week at a United Nations climate meeting in Germany, China and India should easily exceed the targets they set for themselves in the 2015 Paris Agreement... India is now expected to obtain 40 percent of its electricity from non-fossil fuel sources by 2022, eight years ahead of schedule." Solar power in Japan has been expanding since the late 1990s. By the end of 2017, cumulative installed PV capacity reached over 50 GW with nearly 8 GW installed in the year 2017. The country is a leading manufacturer of solar panels and is in the top 4 ranking for countries \\
\textbf{Question}: Which of the following countries generated the most total energy from solar sources in 2019? \\
A. China
B. United States
C. Germany
D. Japan
 \\
\bottomrule
\end{tabularx}
\caption{Prompt for MMLU}
\label{tab:full_instructions1}
\end{table*}

\begin{table*} []
\small
\begin{tabularx}{\textwidth}{m{17cm}}
\toprule
\textbf{Knowledge}: received 122,000 buys (excluding WWE Network views), down from the previous year\'s 199,000 buys. The event is named after the Money In The Bank ladder match, in which multiple wrestlers use ladders to retrieve a briefcase hanging above the ring. The winner is guaranteed a match for the WWE World Heavyweight Championship at a time of their choosing within the next year. On the June 2 episode of "Raw", Alberto Del Rio qualified for the match by defeating Dolph Ziggler. The following week, following Daniel Bryan being stripped of his WWE World Championship due to injury, Stephanie McMahon changed the \\ 
\textbf{Question}: Who won the mens money in the bank match? \\
\textbf{Answer}: Braun Strowman \\
\\
\textbf{Knowledge}: in 3D on March 17, 2017. The first official presentation of the film took place at Disney\'s three-day D23 Expo in August 2015. The world premiere of "Beauty and the Beast" took place at Spencer House in London, England on February 23, 2017; and the film later premiered at the El Capitan Theatre in Hollywood, California, on March 2, 2017. The stream was broadcast onto YouTube. A sing along version of the film released in over 1,200 US theaters nationwide on April 7, 2017. The United Kingdom received the same version on April 21, 2017. The film was re-released in \\ 
\textbf{Question}: When does beaty and the beast take place \\\
\textbf{Answer}: Rococo-era \\
... \\
\textbf{Knowledge}: Love Yourself "Love Yourself" is a song recorded by Canadian singer Justin Bieber for his fourth studio album "Purpose" (2015). The song was released first as a promotional single on November 8, 2015, and later was released as the album\'s third single. It was written by Ed Sheeran, Benny Blanco and Bieber, and produced by Blanco. An acoustic pop song, "Love Yourself" features an electric guitar and a brief flurry of trumpets as its main instrumentation. During the song, Bieber uses a husky tone in the lower registers. Lyrically, the song is a kiss-off to a narcissistic ex-lover who did \\
\textbf{Question}: love yourself by justin bieber is about who \\
 \\
\bottomrule
\end{tabularx}
\caption{Prompt for open-domain QA}
\label{tab:full_instructions1}
\end{table*}




\end{document}